\documentclass[10pt,twocolumn,letterpaper]{article}

\usepackage[review]{iccv}      % To produce the REVIEW version
\usepackage{multirow}
\usepackage{pifont}
\definecolor{iccvblue}{rgb}{0.21,0.49,0.74}
\usepackage[pagebackref,breaklinks,colorlinks,allcolors=iccvblue]{hyperref}

\title{TR-DQ: Time-Rotation Diffusion Quantization}

\author{First Author\\
Institution1\\
Institution1 address\\
{\tt\small firstauthor@i1.org}
\and
Second Author\\
Institution2\\
First line of institution2 address\\
{\tt\small secondauthor@i2.org}
}
\makeatletter
\def\thanks#1{\protected@xdef\@thanks{\@thanks
        \protect\footnotetext{#1}}}
\makeatother

\begin{document}
\maketitle

\begin{abstract}
Diffusion models have been widely adopted in image and video generation. However, their complex network architecture leads to high inference overhead for its generation process. Existing diffusion quantization methods primarily focus on the quantization of the model structure while ignoring the impact of time-steps variation during sampling. At the same time, most current approaches fail to account for significant activations that cannot be eliminated, resulting in substantial performance degradation after quantization. To address these issues, we propose \textbf{Time-Rotation Diffusion Quantization (TR-DQ)}, a novel quantization method incorporating time-step and rotation-based optimization. TR-DQ first divides the sampling process based on time-steps and applies a rotation matrix to smooth activations and weights dynamically. For different time-steps, a dedicated hyperparameter is introduced for adaptive timing modeling, which enables dynamic quantization across different time steps. Additionally, we also explore the compression potential of Classifier-Free Guidance (CFG-wise) to establish a foundation for subsequent work. TR-DQ achieves \textbf{state-of-the-art (SOTA)} performance on image generation and video generation tasks and a 1.38-1.89$\times$ speedup and 1.97-2.58$\times$ memory reduction in inference compared to existing quantization methods. 
\end{abstract}    
\section{Introduction}
\label{sec:intro}

Diffusion models~\cite{ho2022video,xing2024survey} have demonstrated a remarkable ability to generate model parameters~\cite{shao2025context}, 3D scenes~\cite{erkocc2023hyperdiffusion,wang2024occsora, yan20243dsceneeditor}, etc. Also, they outperform GANs~\cite{goodfellow2014generative,goodfellow2020generative} in most image and video generation tasks. However, due to their high memory consumption during inference, diffusion models are challenging to deploy on edge devices. In addition, the generation process consumes significant latency at each time-step, leading to low throughput, particularly for high-resolution images and long video generation. Therefore, compressing diffusion models while preserving their generative capability is crucial for practical deployment.

Several model compression methods are currently being tested in diffusion. In most of model compression methods~\cite{buciluǎ2006model,cheng2017survey,zhu2024survey}. Quantization offers a promising solution to reduce memory and speed up computation for deployment on limited-resource devices. However, among them, post-training quantization (PTQ) methods~\cite{frantar2022gptq,lin2024awq,dettmers2023spqr} could avoid retraining the model, but do not achieve satisfactory results when applied directly to diffusion models. The distribution of time-steps significantly influences the generation process; hence, ignoring the activation distribution at each time-step can lead to negative effects. To address these issues, Q-Diffusion~\cite{li2023q} introduces a timestep-aware calibration and realizes end-to-end quantization, which enables the quantization of full-precision unconditional diffusion models into 4-bit. QUEST~\cite{watson1983quest} identifies three key properties in quantized diffusion models affecting current methods: imbalanced activation distributions, imprecise temporal information, and specific module perturbation vulnerability.

However, most of these methods mentioned above ignore the effect of significance activation in the diffusion model and therefore cause additional losses in quantification. In addition, Classifier-Free Guidance (CFG) is also a major factor that is ignored~\cite{xie2024litevar}. To address these problems, we propose a time-step and rotation based quantization method, \textbf{Time-Rotation Diffusion Quantization (TR-DQ)}. TR-DQ first transfers the massive activations into weights using a rotation metrix, which makes activations and weights smoother and easier to quantize. Meanwhile, we explore both Classifier-Free Guidance (CFG) and non-CFG based quantization. Notably, we observe that some layers in CFG and non-CFG share similar parameter sensitivity distributions, allowing us to further compress them through a merging-based approach.

In order to demonstrate the effectiveness of our methodology, we conducted extensive experiments on image generation and video generation tasks. Experimental results demonstrate that our method outperforms existing quantization techniques in both image and video generation across most metrics, achieving state-of-the-art (SOTA) performance. Meanwhile, our method can achieve 1.7$\times$ speedup, significantly enhancing the efficiency of the generative model. The key contributions of our work are as follows:
\begin{itemize}
    \item We shift the hard-to-quantify activations into weights using a rotation matrix, resulting in a smoother activation distribution that is easier to quantize.
    
    \item We introduce a novel quantization approach, Time-Rotation Diffusion Quantization (TR-DQ), by extending the global rotation matrix into a time-dependent rotation matrix based on the time-step distribution of diffusion models.

    \item By analyzing the similarity in attentional sensitivity between CFG and non-CFG, we implement attentional merging quantization to optimize compression.

    \item Our method significantly reduces quantization loss while preserving high visual quality in image and video generation tasks while achieving 1.38-1.89$\times$ speedup and 1.97-2.58$\times$ memory reduction without compromising performance..
\end{itemize}

\section{Related Work}
\begin{figure*}
    \centering
    \includegraphics[width=1\linewidth]{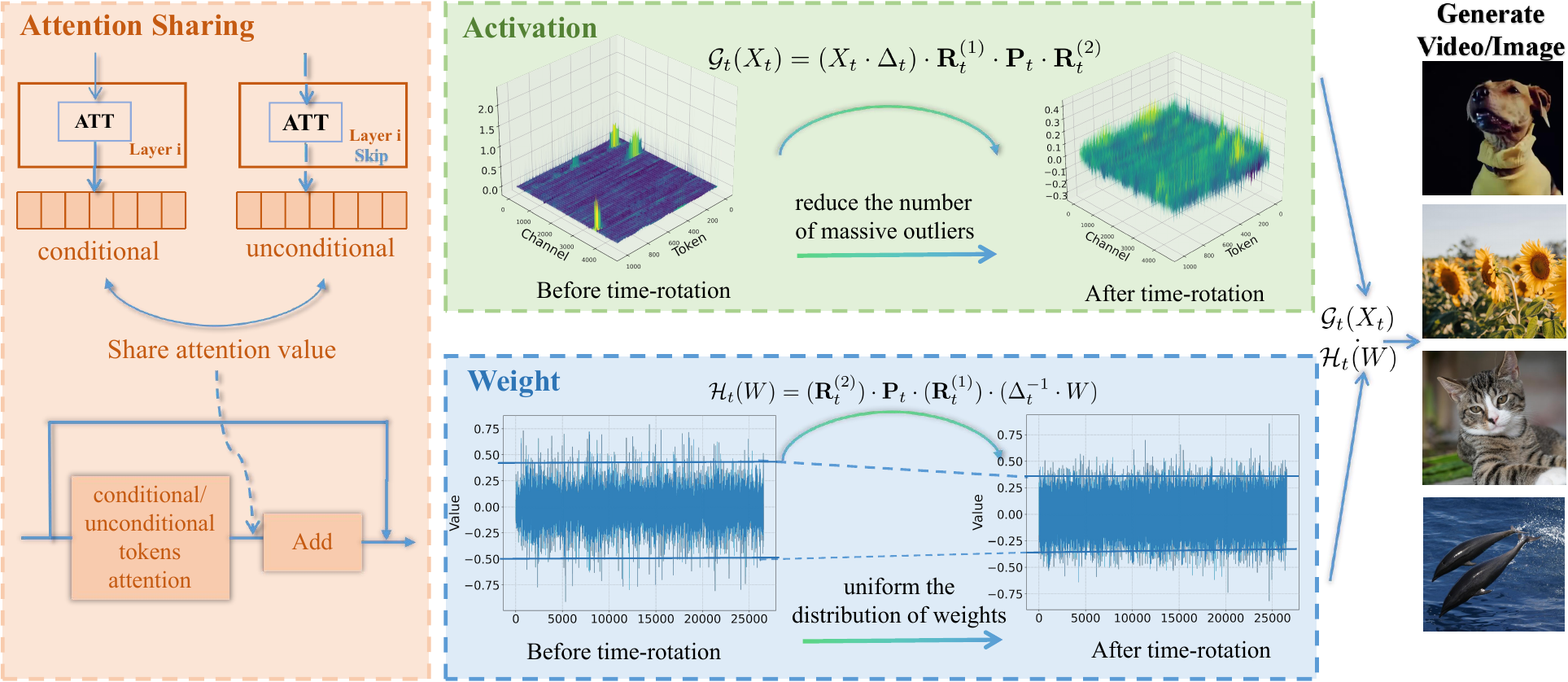}
    \caption{\textbf{Main pipeline of TR-DQ. }TR-DQ uses a rotation matrix for the activations to reduce the massive outliers, and also rearranges the weights to be a smoother and easier to quantify model overall. For CFG and non-CFG with high similarity of attention TR-DQ performs weight sharing, which further reduces the computational cost.}
    \label{pipeline}
\end{figure*}
\subsection{Generative Models}
Image and video generation has achieved remarkable progress. Early GAN-based video generation methods~\cite{gupta2022rv,liu2021generative} have temporal coherence problems and consecutive frame discrepancies.  Similarly, GAN-based image generation models~\cite{karras2019style} are known for their instability during training, frequently encountering problems such as mode collapseand. VAE-based methods~\cite{yan2021videogpt,kingma2013auto, yan2024renderworld} provide a robust framework but often require extensive computational resources. Video diffusion models with U-Net architecture were adapted to boost frame continuity. Latte~\cite{ma2024latte} pioneered the use of transformer~\cite{vaswani2017attention} to realize high quality text-to-video generation, outperforming traditional methods in processing complex video data. SORA~\cite{brooks2024video,zheng2024open} further inspired the development of video diffusion transformer, advancing the development of models like GenTron~\cite{chen2024gentron}, which extended the capabilities of diffusion transformers to multi-frame video generation. Tora~\cite{jankovic1996tora} focuses on trajectory-oriented video generation and combines textual, visual, and trajectory conditions to create high-quality videos. However, existing image and video generation models still suffer from high memory cost. To address this issue, approaches such as model quantization, pruning and distillation are proposed. In our work, we mainly focus on model quantization, and we will review milestones of diffusion quantization in Section \ref{sec2.2}.

\subsection{Diffusion Model Quantization}
\label{sec2.2}
The evolution of model quantization has been instrumental in enabling the deployment of complex neural networks on resource-constrained devices. Post-training quantization (PTQ) methods like RTN~\cite{nagel2020up} and LLM.int8~\cite{dettmers2022gpt3} quantize weights and activations post-training with a few calibration dataset. However, most of the quantization methods are not suitable for diffusion models because diffusion models contain time-steps with different activation each steps. To address this, Q-Diffusion~\cite{li2023q} proposes a PTQ method for diffusion models, compressing them to 4-bit without performance loss by time-step-aware sampling and separation shortcut quantization. PTQ4DM~\cite{shang2023post} uses time-step-aware and separation shortcut techniques to compress models to 4-bit with similar performance to full-precision ones, and SVDQuant~\cite{li2024svdqunat} quantizes diffusion model weights and activations to 4-bit by introducing a low-rank branch to absorb outliers. Q-DiT~\cite{chen2024q} customizes quantization parameters for channels to address imbalance, while PTQ4DiT~\cite{wu2024ptq4dit} has designed a fixed mask adaptable to all timesteps to handle time-varying imbalance. For video generation model, ViDiT-Q~\cite{zhao2024vidit} designs a post-training quantization (PTQ) method for DiTs that enables W8A8 lossless quantization and W4A8 quantization without loss of generation quality. In our work, we mainly focus on time-steps modeling and explore the impact of Classifier-Free Guidance (CFG) wise.

\section{Methodology}

\subsection{Preliminaries}
As blocks of diffusion models are predominantly constructed with basic linear layers, which can be represented as, $Y = X\cdot W$. Here, $W$ is the weight matrix. $X$ and $Y$ are denoted as input activations and output activations, respectively. In this paper, we focus on integer uniform quantization of both activations and weights, aiming to achieve better hardware support. Specifically, the $b$-bit quantization process maps the FP16 tensor $X$ to low-bit integer $X_{int}$ could be expressed as Eq.~\eqref{Q1},
% \begin{equation}
% X_{int} = clamp(\lfloor \frac{X}{s} \rceil + z, 0, 2^b-1),
% \label{Q1}
% \end{equation}
\begin{equation}
    X_{\textbf{int}} = \operatorname{clamp} \left( \left\lfloor \dfrac{X}{s} \right\rceil + z, 0, 2^b - 1 \right),
\label{Q1}
\end{equation}
where the function $clamp(x,0,2^b-1)$ clamps the values $x$ into range $[0, 2^b-1]$, the nation $\lfloor \ \cdot \rceil $ means the nearest rounding operation. The scaling $s$ could be expressed as Eq~\eqref{Q2},
% \begin{equation}
%      s = (max(X)-min(X))/(2^b-1),
%      \label{Q2}
% \end{equation}
\begin{equation}
    s = \frac{\max(X) - \min(X)}{2^b - 1}.
\label{Q2}
\end{equation}
and the zero point $z$ could be calculated as Eq~\eqref{Q3},
% \begin{equation}
%     z = \lfloor min(X)/s \rceil,
%     \label{Q3}
% \end{equation}
\begin{equation}
    z = \left\lfloor \frac{\min(X)}{s} \right\rceil.
\label{Q3}
\end{equation}
  
Following recent work, we employ per-token quantization for activations and per-channel quantization for weights.

For diffusion models, the presence of outliers in activations poses significant challenges to activations quantization. To address this issue, current quantization methods like SmoothQuant~\cite{xiao2023smoothquant}  typically employ smoothing techniques, using computational invariance to shift the quantization difficulty from activations to weights. It's formula is as Eq.~\eqref{Q4},
% \begin{equation}
% Y = (Xdiag(\Delta)^{-1})(diag(\Delta)W) = \hat{X}\cdot \hat{W},
% \label{Q4}
% \end{equation}
\begin{equation}
    Y = \big(X \operatorname{diag}(\Delta)^{-1} \big) 
    \big( \operatorname{diag}(\Delta) W \big) 
    = \hat{X} \cdot \hat{W}.
\label{Q4}
\end{equation}

The diagonal element $\Delta_j$ within $\Delta$ is computed as Eq.~\eqref{Q5},
% \begin{equation}
%     \Delta_j = \max(|X_j|)^{\alpha}/\max(|W_j|)^{1-\alpha}),
%     \label{Q5}
% \end{equation}
\begin{equation}
    \Delta_j = \frac{\operatorname{max} \big( \lvert X_j \rvert \big)^{\alpha} }
    {\operatorname{max} \big( \lvert W_j \rvert \big)^{1-\alpha}},
\label{Q5}
\end{equation}
where $\alpha$ is a hyper-parameter representing the migration strength. 

\subsection{Quantization Strategies}

\begin{figure}[h]
    \centering
    \begin{subfigure}{0.48\linewidth}
        \centering
        \includegraphics[width=\linewidth]{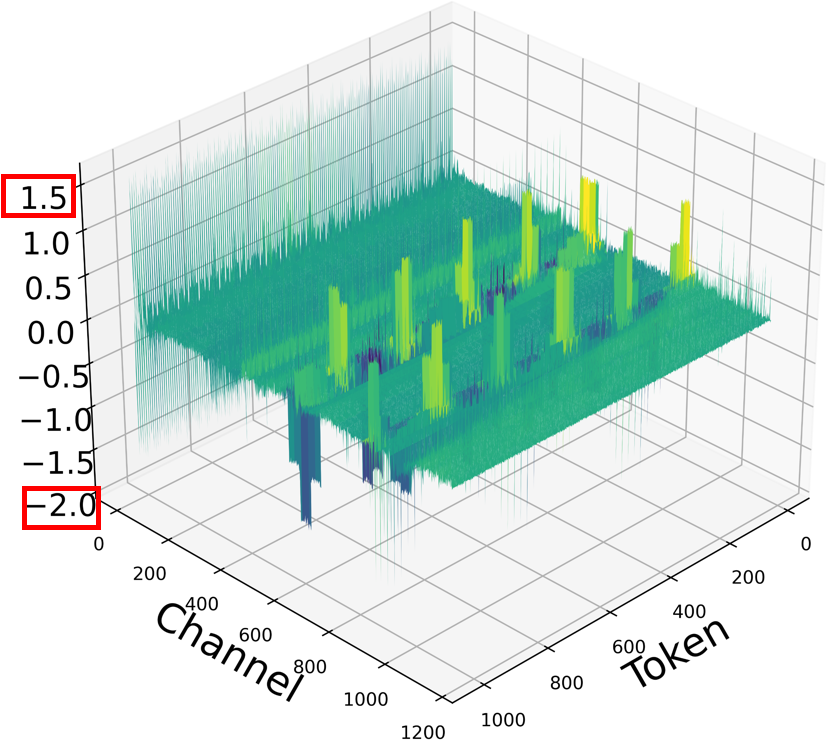}
        \captionsetup{font={footnotesize,bf,stretch=1},justification=raggedright}
        \caption{X without Time-Rotation.}
        \label{fig:subfig1a}
    \end{subfigure}
    \hfill
    \begin{subfigure}{0.48\linewidth}
        \centering
        \includegraphics[width=\linewidth]{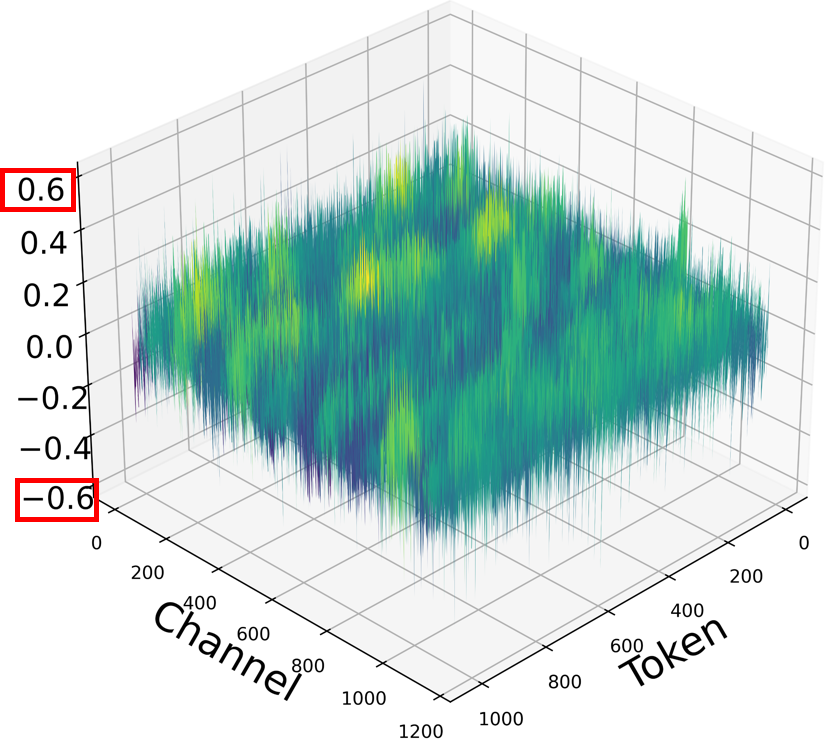}
        \captionsetup{font={footnotesize,bf,stretch=1},justification=raggedright}
        \caption{X with Time-Rotation.}
        \label{fig:subfig1b}
    \end{subfigure}
    
    \medskip  
    
    \begin{subfigure}{0.48\linewidth}
        \centering
        \includegraphics[width=\linewidth]{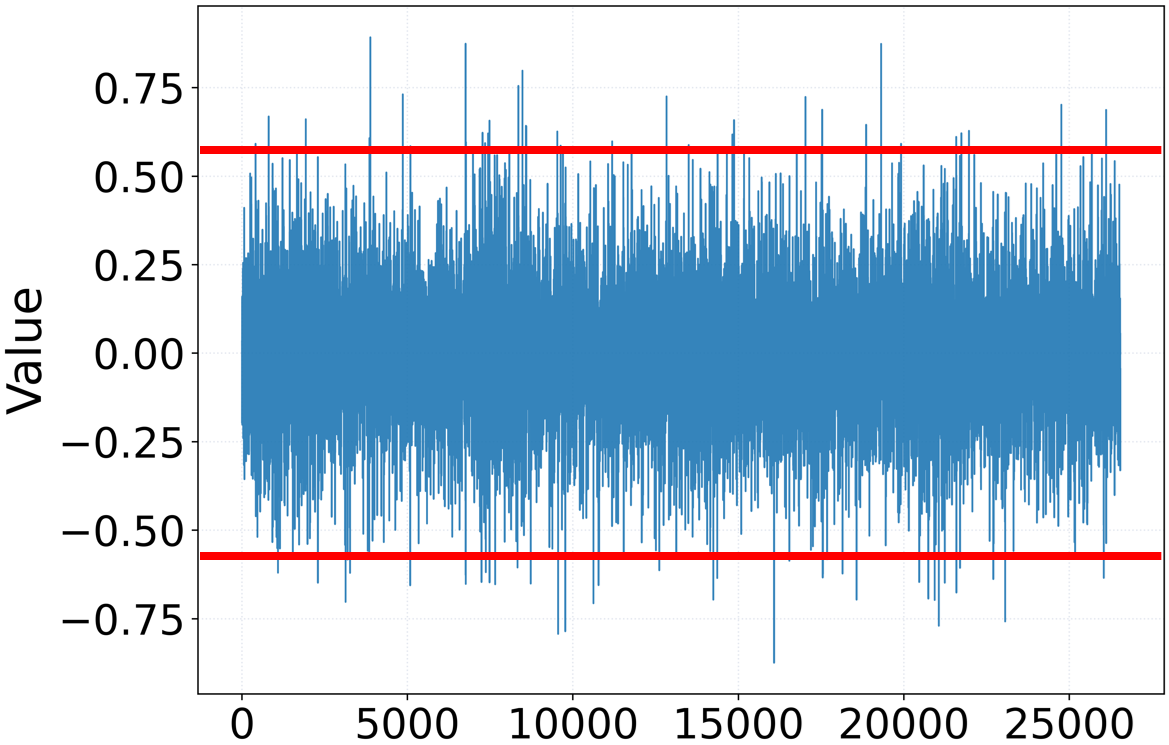}
         \captionsetup{font={footnotesize,bf,stretch=1},justification=raggedright}
        \caption{W without Time-Rotation.}
        \label{fig:subfig1c}
    \end{subfigure}
    \hfill
    \begin{subfigure}{0.48\linewidth}
        \centering
        \includegraphics[width=\linewidth]{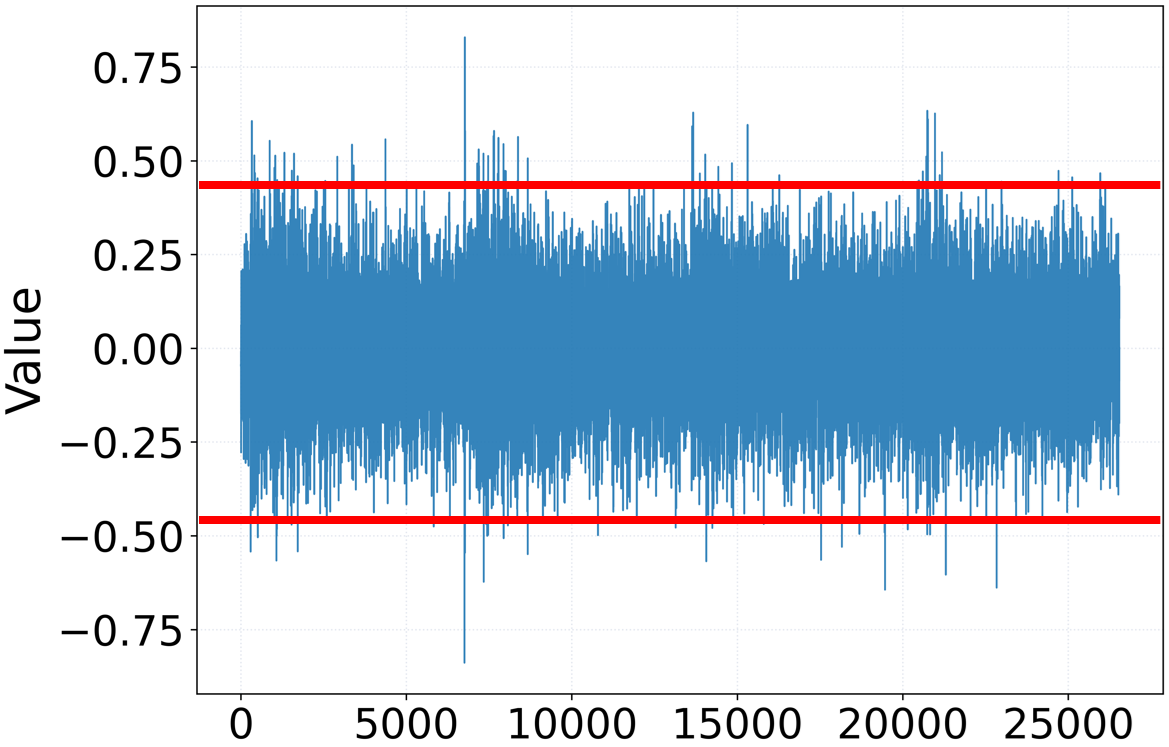}
         \captionsetup{font={footnotesize,bf,stretch=1},justification=raggedright}
        \caption{W with Time-Rotation.}
        \label{fig:subfig1d}
    \end{subfigure}
    \caption{\textbf{Effect of Time-Rotation on Data Distribution. }Data distribution with \textbf{Time-Rotation} is more smoother. Where $X$ is the activations and $W$ is the weights.}
    \label{fig:Figure1}
\end{figure}

Examining~\Cref{fig:subfig1a} reveals that although smooth techniques reduce some outliers in activations, certain difficult to smooth outliers which we term \textbf{Massive Outliers} still persist. Although some outliers were smoothed, this did not change the unevenness of the data distribution. All these factors affect quantization performance. Additionally, since the quantization difficulty is transferred from activations to weights, the weight distribution becomes even more irregular, making weight quantization another challenge \cite{shao2024gwq}.
Therefore, adopting a novel balancing strategy to equilibrate activations and weights is necessary. As it shown in Fig.~\ref{pipeline}, we leverage rotation matrices based on computational invariance. Through rotation matrices, we can reduce the number of \textbf{Massive Outliers} in activations and make the data distribution of both activations and weights more uniform, facilitating group-wise quantization. The specific details are as follows:

\noindent \textbf{Balancing Strategies.}
Based on these observations and building upon DuQuant~\cite{lin2025duquant}, we utilize an orthogonal rotation matrix R, a matrix constructed based on prior knowledge and greedy strategies, which can identify and swap the positions of outliers The construction of this rotation matrix is as Eq~\eqref{Q6},
\begin{equation}
    \mathbf{R^1}=\mathbf{C_1}{\mathbf{Q} \mathbf{C_2}}, \label{Q6}
\end{equation}
where the $\mathbf{C_1}$ is the switching matrix used to swap the first column and the column containing the maximum outlier columns of the activations,
and $\mathbf{Q}$ represents an orthogonal randomly initialized rotation matrix, in which the first row is specifically uniformly distributed. The motivation behind this is to mitigate outliers in the first column after the transformation by $\mathbf{C_1}$. To preserve the structure of the activations, we use $\mathbf{C_2}$ to perform the inverse operation of $\mathbf{C_1}$, specifically swapping those two columns back again.

Thus, we obtain the final rotation matrix through a greedy strategy, with the formula as Eq.~\eqref{Q7},
\begin{equation}
     \mathbf{R} = \mathbf{R}^1\mathbf{R}^2\cdots \mathbf{R}^n,
     \label{Q7}
\end{equation}
where $n = \arg\min_{k\in[1:N]} \left(max_{i,j} |(\mathbf{X} \mathbf{R}^1\cdots \mathbf{R}^k)_{ij}|\right)$. Each $\mathbf{R}^i$ is constructed according to~\cref{Q6}.
Through this construction manner, we can ensure that the final rotation matrix ${\mathbf{R}}$ can effectively mitigate outliers with large magnitudes, as opposed to merely using a randomly selected orthogonal rotation matrix. Nevertheless, directly constructing the entire rotation matrix is time-consuming and results in substantial memory overhead. For fast matrix multiplication, following~\cite{lin2025duquant}, we approximate the rotation matrix $\mathbf{R} \in \mathbb{R}^{C_{in} \times C_{in}}$ in a block-wise manner:
\begin{equation}
    \mathbf{R} =~\textbf{BlockDiag}(\mathbf{R}_{b_1}, ..., {\mathbf{R}}_{b_K}), \label{Q4}
\end{equation}
where $\mathbf{R}_{b_i} \in \mathbb{R}^{2^n \times 2^n}$ denotes a square matrix of the $i$-th block, which is constructed following the three steps mentioned above. And the block numbers $K$ is calculated by $K = C_{in}/{2^n}$.
After the first block rotation, most \textbf{Massive Outliers} can be eliminated, but in order to make the data smoother for per-token quantization, we need to perform a second block rotation. However, When we using the first block rotation reduces outliers locally, the distribution between different blocks may remain imbalanced, which is unfavorable for our second block rotation. To address this issue, we introduce the \textbf{zigzag  permutation}. Concretely, we generate a zigzag sequence that starts by assigning channels with the highest activations to the first block. The process continues by assigning channels with the next highest activations to the subsequent blocks in descending order until the end of block $K$. Upon reaching the final block, the order reverses, starting from the channel with the next highest activations and proceeding in ascending order. This back-and-forth patterning continues throughout all the blocks, ensuring that no single block consistently receives either the highest or lowest activations channels. It is worth noting that the constructed permutation is also an orthogonal matrix, which we denote as $\mathbf{P}$. By employing the zigzag permutation, we achieve a balanced distribution of outliers across different blocks. This allows us to use the second block rotation to further smooth the outliers. The final balancing strategy can be represented as Eq.~\eqref{Q9},
% \begin{equation}
% \begin{aligned}
%     Y = X\cdot W=[(X\cdot\Delta)\mathbf{R^{(1)}}\cdot \mathbf{P}\cdot \mathbf{R^{(2)}}]\cdot
%     \\ [\mathbf{(R^{(2)})^\top}\cdot \mathbf{P^\top} \cdot (\mathbf{R^{(1)})^\top}(\Delta^{-1}\cdot W)]\label{Q9}
% \end{aligned}
% \end{equation}
\begin{equation}
\begin{aligned}
    Y &= X\cdot W \\
      &= \big[(X\cdot\Delta)\mathbf{R^{(1)}}\cdot \mathbf{P}\cdot \mathbf{R^{(2)}}\big] \cdot \\
      &\quad \big[\mathbf{(R^{(2)})^\top} \cdot \mathbf{P^\top} \cdot (\mathbf{R^{(1)})^\top} (\Delta^{-1}\cdot W)\big],
\end{aligned}
\label{Q9}
\end{equation}
where the notation $\mathbf{P}$ denotes the orthogonal permutation matrix learned via the zigzag manner, the $\mathbf{R^{(1)}}$ and $\mathbf{R^{(2)}}$ represent the first and second block-diagonal rotation matrix, respectively. Through the application of the second rotation matrix, the activations values become smoother.

\noindent \textbf{Time-Steps Awareness Quantization.}
Since activations at each time-step in diffusion models are different, applying a set of static quantization parameters to these activations would severely damage the generation quality of diffusion models. Furthermore, as activations at each time-step in diffusion models vary, the distribution of outliers in activations across different time-steps also differs significantly. If we still use a single set of $\mathbf{R}$, $\mathbf{P}$ and $\Delta$ for activations across all time-steps, this would ignore the distinctive characteristics of diffusion models and similarly impair their generation quality. To address these issues, we have implemented two approaches. First, we implement dynamic quantization for activations, calculating quantization parameters online. This process only requires additional computation of maximum and minimum values, making the computational cost negligible. Second, based on the characteristics of diffusion models, we propose \textbf{Time-Rotation}, which models the relationship between time and rotation matrices etc. Throughout the denoising process, instead of sharing a single set of $\mathbf{R}$, $\mathbf{P}$ and $\Delta$, activations will select appropriate of $\mathbf{R}$, $\mathbf{P}$ and $\Delta$ based on the current time-step. Therefore, our final formula is as follow:
% \begin{equation}
% \begin{aligned}
% \label{Q10}
% &F_t(X_t,W) = \mathcal{G}_t(X_t) \cdot \mathcal{H}_t(W),
% \\&\mathcal{G}_t(X_t) = (X_t \cdot \Delta_t)\cdot\mathbf{R}^{(1)}_t\cdot\mathbf{P}_t \cdot \mathbf{R}^{(2)}_t,
% \\&\mathcal{H}_t(W) = (\mathbf{R}^{(2)}_t)^\top \cdot \mathbf{P}^\top_t \cdot (\mathbf{R}^{(1)}_t)^\top \cdot(\Delta_t^{-1} \cdot W),
% \\&F_t:\mathbb{R}^{T\times C_{in}}\times\mathbb{R}^{C_{in}\times C_{out}}\longrightarrow \mathbb{R}^{T\times C_{out}},
% \\&t\in T=\{1,2,...,N\}.
% \end{aligned}
% \end{equation}

\begin{equation}
\begin{aligned}
    \label{Q10}
    &F_t(X_t,W) = \mathcal{G}_t(X_t) \cdot \mathcal{H}_t(W), \\
    &\quad \mathcal{G}_t(X_t) = \left( X_t \cdot \Delta_t \right) \cdot \mathbf{R}^{(1)}_t \cdot \mathbf{P}_t \cdot \mathbf{R}^{(2)}_t, \\
    &\quad \mathcal{H}_t(W) = (\mathbf{R}^{(2)}_t)^\top \cdot \mathbf{P}^\top_t \cdot (\mathbf{R}^{(1)}_t)^\top \cdot (\Delta_t^{-1} \cdot W), \\
    &\quad F_t: \mathbb{R}^{T\times C_{in}} \times \mathbb{R}^{C_{in} \times C_{out}} \to \mathbb{R}^{T\times C_{out}}, \\
    &\quad t \in T = \{1,2,\dots,N\}. 
\end{aligned}
\end{equation}

Compared to ViDiTQ~\cite{zhao2024vidit}'s Quarot ~\cite{ashkboos2024quarot} based rotation matrices, our \textbf{Time-Rotation} are more sophisticated in construction. Rather than being simply initialized, they are constructed based on prior knowledge and greedy strategies. In addition, the extra permutation operation enables better handling of unevenly distributed data. As demonstrated in \Cref{fig:subfig1b} and \Cref{fig:subfig1d}, this method effectively smooths the distribution of both activations and weights, with the activations values range narrowing from [-2.0, 1.5] to [-0.6, 0.6], and the weight values become smoother. This evidence confirms that our method achieves superior smoothing effects compared to conventional smoothing approaches.

\begin{figure}[h]
    \centering
    \begin{subfigure}{0.48\linewidth}
        \centering
        \includegraphics[width=\linewidth]{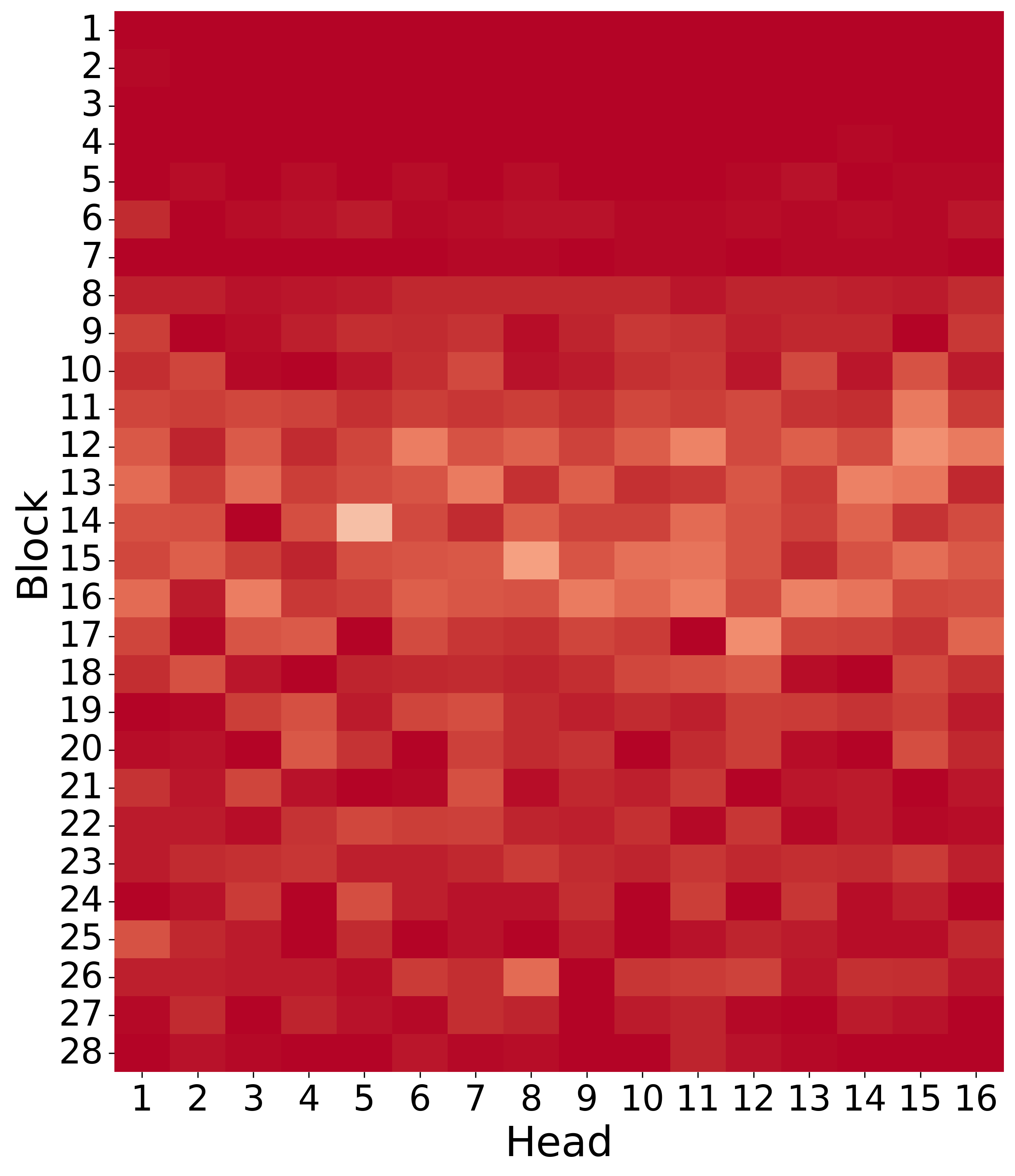}
        \caption{Heat map at time-step=12.}
        \label{fig:subfig3a}
    \end{subfigure}
    \hfill
    \begin{subfigure}{0.48\linewidth}
        \centering
        \includegraphics[width=\linewidth]{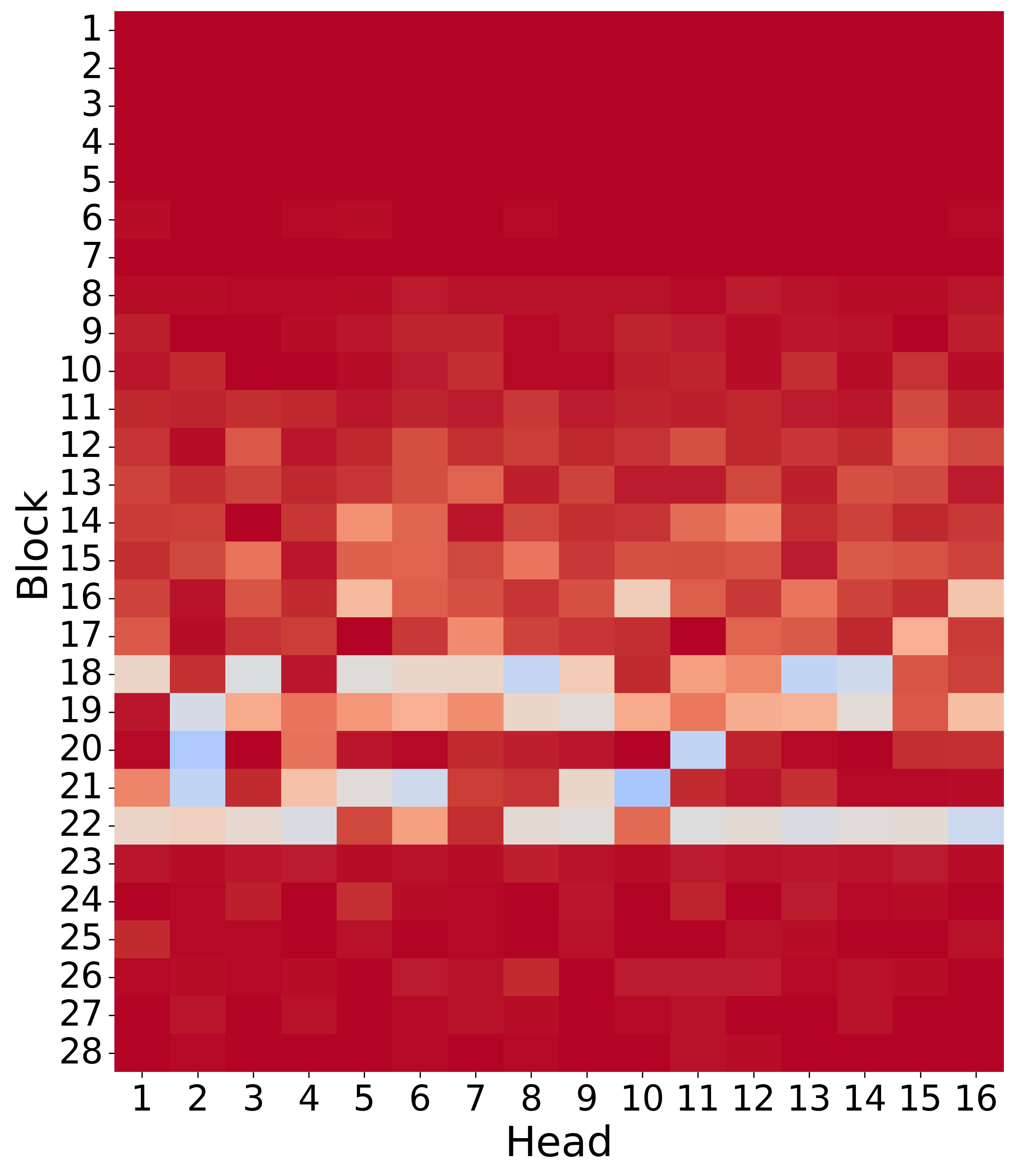}
        \caption{Heat map at time-step=15.}
        \label{fig:subfig3b}
    \end{subfigure}
    \caption{\textbf{Heat maps of multi-head self-attention under conditional and unconditional situations.}  Each square reflects the similarity between the two.  The redder the square, the
    higher the similarity;  the bluer the square, the lower the similarity.}
    \label{}
\end{figure}

\subsection{Attention-Sharing Quantization (AS)}
Classifier-free guidance (CFG) is widely used for diffusion transformers, enabling the generation of more imaginative images or videos that are not confined to a single format. However, the adoption of CFG technology means that we cannot complete the task with just a single denoising process, which significantly slows down inference speed. To address this issue, through examination of the structure of diffusion transformers, we discovered that there exists substantial similarity between attention values of multi-head self-attention in each block of the condition and unconditional paths, as shown in~\Cref{fig:subfig3a}. Our evaluation metric is cosine similarity, detailed as follows Eq.~\eqref{Q11},
% \begin{equation}
% \cos(\mathbf{A},\mathbf{B}) = \frac{\mathbf{A} \cdot \mathbf{B}}{\|\mathbf{A}\| \|\mathbf{B}\|}
% \label{Q11}
% \end{equation}
\begin{equation}
    \cos(\mathbf{A},\mathbf{B}) = \frac{\mathbf{A} \cdot \mathbf{B}}
    {\left\|\mathbf{A}\right\| \left\|\mathbf{B}\right\|}.
\label{Q11}
\end{equation}

However, due to the time-step differences mentioned earlier, not all attention values are identical across every time-step, as illustrated in~\Cref{fig:subfig3b}. Through extensive observation, we found that the attention values of multi-head self-attention in the forward and backward blocks show significant similarities across all time-steps. Therefore, we decided to implement Attention-Sharing for these blocks. Subsequent experiments demonstrated that our Attention-Sharing approach improves inference speed without compromising the quality of generated images or videos.
\section{Experiment}
\begin{table*}[htbp]

\centering
    \begin{tabular}{ccccccccccc}
\toprule
\multirow{2}{*}{\textbf{Method}} & \textbf{Bit-width} & Imaging & Aesthetic & Motion & Dynamic & BG. & Subject & Scene & Overall \\
& (W/A) & Quality & Quality & Smooth. & Degree &  Consist. & Consist. & Consist. & Consist. \\ 
\midrule 
 - & 16/16 & 63.68 & 57.12 & 97.01 & 56.94 & 96.13 & 92.28 & 40.51 & 26.21 \\
\midrule 
 % Naive PTQ & 8/8 & 56.52 & 53.64 & 95.85 & 69.44 & 93.54 & 85.30 & 29.28 & 25.42 &   \\
 Q-Diffusion~\cite{li2023q} & 8/8 & 60.38 & 55.15 & 94.44 & 68.05 & 94.17 & 87.74 & 36.62 & 25.66 &   \\

 Q-DiT~\cite{chen2024q} & 8/8 & 60.35 & 55.80 & 93.64 & 68.05 & 94.70 & 86.94 & 32.34 & 26.09 &   \\
 PTQ4DiT~\cite{wu2024ptq4dit} & 8/8 & 56.88 & 55.53 & 95.89 & 63.88 & 96.02 & 91.26 & 34.52 & 25.32 &  \\
 % SQ-Static & 8/8 & 61.74 & 55.93 & 95.79 & 66.66 & 95.18 & 88.70     & 35.73 & 26.31 &   \\

 ViDiT-Q~\cite{zhao2024vidit} & 8/8 & 61.48 & 56.95 & 96.14 & 61.11 & 95.84 & 90.24 & 38.22 & 26.06 &   \\

 TR-DQ & 8/8 & 61.82 & 57.44 & 96.63 & 55.38 & 96.11 & 91.14 & 39.78 & 26.18    \\
 TR-DQ+AS & 8/8 & 60.38 & 57.10 & 96.26 & 50.27 & 95.71 & 91.58 & 38.50 & 25.99    \\
\midrule 
  % & ViDiT-Q-MP & 6/6 & 62.07 & 57.03 & 95.86 & 62.50 & 95.86 & 89.34 & 39.46 & 26.41   \\

Q-DiT~\cite{chen2024q} & 4/8 & 23.30 & 29.61 & 97.89 & 4.166 & 97.02 & 91.51 & 0.00 & 4.985 &   \\
PTQ4DiT~\cite{wu2024ptq4dit} & 4/8 & 37.97 & 31.15 & 92.56 & 9.722 & 98.18 & 93.59 & 3.561 & 11.46 &  \\
 
 % ViDiT-Q & 4/8 & 49.25 & 45.48 & 98.20 & 27.77 & 96.91 & 91.46 & 25.41 & 20.42 \\

 ViDiT-Q\cite{zhao2024vidit} & 4/8 & 59.01 & 55.37 & 95.69 & 48.33 & 95.23 & 88.72 & 36.19 & 25.94 &\\

TR-DQ & 4/8 & 59.88 & 56.20 & 96.57 & 51.83 & 96.65 & 90.74 & 32.46 & 26.17   \\
TR-DQ+AS & 4/8 & 57.69 & 55.02 & 96.74 & 47.78 & 96.58 & 91.03 & 32.25 & 25.34    \\
\bottomrule
\vspace{-20pt}
\end{tabular}
\caption{\textbf{Performance of TR-DQ on video generation on VBench.} TR-DQ outperforms the current SOTA ViDiTQ in most metrics, suggesting that it is more capable of generating models after quantization.}
\label{tab:main_vbench}
\end{table*}

\begin{table}[]
    \centering
    \resizebox{1\linewidth}{!}{
     \begin{tabular}{ccccc}\toprule
         \multirow{2}{*}{\textbf{Method}} & \textbf{Bit-width} & \multirow{2}{*}{\textbf{FID($\downarrow$)}} & \multirow{2}{*}{\textbf{CLIP($\uparrow$)}} & \multirow{2}{*}{\textbf{IR($\uparrow$)}} \\
             & (W/A) & & & \\ \midrule
         
            % \multirow{9}{*}{\textbf{\large Pixart-$\alpha$}} & 
            - & 16/16 &  73.34 &  0.258 &  0.901 \\ \midrule
            Q-Diffusion~\cite{li2023q}  & 8/8 & 96.54 & 0.239 & 0.186\\
            Q-DiT~\cite{chen2024q}&  8/8 &  73.60 &  0.256 &  0.854\\
            PTQ4DiT~\cite{wu2024ptq4dit} &  8/8 &  127.9 &  0.217 &  -1.216 \\
            ViDiT-Q~\cite{zhao2024vidit} &  8/8 &  75.98 &  0.232 &  0.859\\
            TR-DQ &  8/8 &  75.12 &  0.249 &  0.887\\ 
            TR-DQ+AS&  8/8 &  75.57 &  0.233 &  0.863\\
            \midrule
            
            Q-Diffusion~\cite{li2023q}   & 4/8 & 91.95& 0.228 & -0.224\\
            Q-DiT~\cite{chen2024q} & 4/8 & 475.8 & 0.127 & -2.277\\
            PTQ4DiT~\cite{wu2024ptq4dit}& 4/8 & 171.9 & 0.177 & -2.064\\
            ViDiT-Q~\cite{zhao2024vidit} & 4/8 & 76.65 & 0.243 & 0.837\\
            TR-DQ  & 4/8 & 75.53 & 0.252 & 0.851\\
            TR-DQ+AS  & 4/8 & 75.76 & 0.246 & 0.847\\
            % & \multirow{2}{*}{Naive}  & 8/8 & 115.14 & 0.226 & -0.953\\
            % & & 4/8 & 108.40 & 0.234 & -0.725\\
            %  \cmidrule(lr){2-6}
            
            \bottomrule
                \vspace{-20pt}
    \end{tabular}}

    \caption{\textbf{Results of image generation task. }TR-DQ method has an overall advantage over current quantization methods for the same bits. AS indicates Attention-Sharing. In addition, it is worth noting that \textbf{ViDiT-Q's W4A8 uses a mixed quantization} that means the weights are not really 4-bit quantization, there may be 6 and 8 bits.}
    \label{tab:image}
\end{table}
\subsection{Main Results}
\noindent\textbf{Image Generation Tasks.}
\begin{figure}
    \centering
    \includegraphics[width=\linewidth]{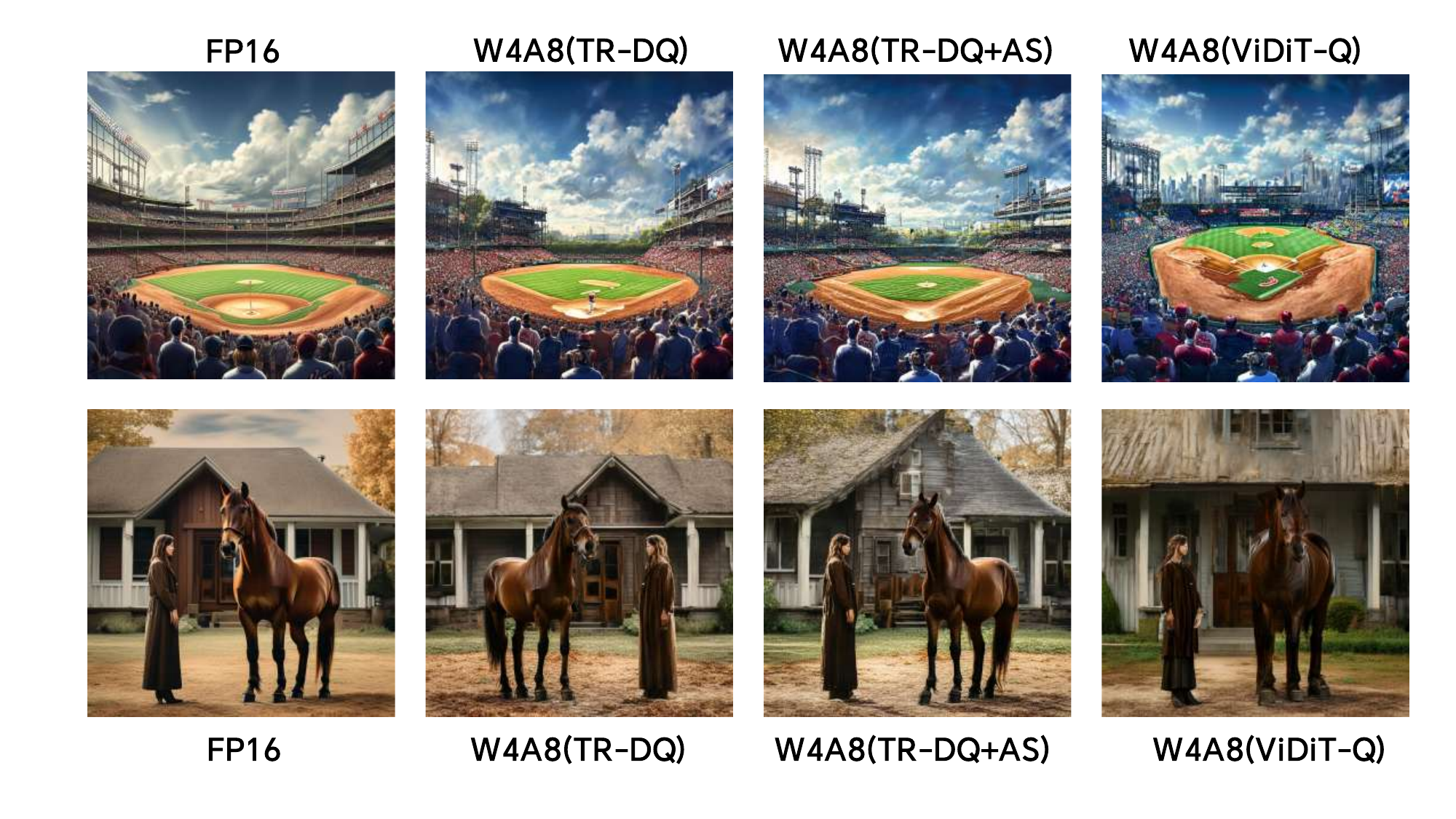}

    \caption{\textbf{Visualisation of image.} TR-DQ and the model adding weight sharing generated better quality images.}
    \label{image}
\end{figure}
In this section, we will discuss the superiority of our method through the existing benchmark and visualization results. To ensure comprehensiveness, we will introduce some of the existing quantization methods for large language models (LLMs) as baseline for comparison.

\subsection{Experiment Setting}
\textbf{Hardware Setting.}
We follow DuQuant's~\cite{lin2025duquant} hardware design scheme to transfer the rotation matrix computation to CUDA to improve the quantization efficiency. During the experiments, the CUDA version was always 12.1. 

\noindent\textbf{Setting up.}
For the fairness of the experiment, we follow the experiment setting up of VIDITQ~\cite{zhao2024vidit}. 
For the image generation task, we use PixArt-$\alpha$~\cite{chen2023pixart} pre-trained on COCO dataset~\cite{lin2014microsoft} for evaluation. During the image generation process, we set the STEP to 20 and the CFG scale to 4.5. For the image generation task, we use Open-SORA~\cite{zheng2024open} pre-trained on UCF-101~\cite{soomro2012ucf101} for evaluation. During the video generation process, we also set the STEP to 20 and the CFG scale to 4.5. All experience were done with GPU NVIDIA A800 (80G).

\begin{table*}[]
   \centering
        \resizebox{0.75\linewidth}{!}{
            \begin{tabular}{ccccccc}
            \toprule
            % \multirow{2}{*}{\textbf{Model}} & 
            \multirow{2}{*}{\textbf{Method}} & \textbf{Bit-width} & \multirow{2}{*}{\textbf{CLIPSIM}} &  \multirow{2}{*}{\textbf{CLIP-Temp}} & VQA- & VQA- & $\Delta$ Flow  \\
             & (W/A) & & & \small{Aesthetic} & \small{Technical} & Score. ($\downarrow$) \\
            \midrule
            
            % \multirow{24}{*}{\textbf{\large STDiT}} & 
            - & 16/16 &  0.1818 & 0.9988 & 63.40 & 50.46 & - \\
             \cmidrule(lr){1-7}
             % & Naive PTQ & 8/8 & 0.1956 & 0.9988 & 48.2358 & 25.5961 & 0.367  \\
             % PTQ4DM & 8/8 & 0.1812 & 0.9984 & 50.0674 & 25.1344 & 0.342  \\
             Q-Diffusion~\cite{li2023q} & 8/8 & 0.1781 & 0.9987 & 51.68 & 38.27 & 0.328 \\
             % PTQD & 8/8 & 0.1778 & 0.9981 & 51.8039 & 36.6078 & 0.031 \\
             Q-DiT~\cite{chen2024q} & 8/8 & 0.1788 & 0.9977 & 61.03 & 34.97 & 0.473 \\
             PTQ4DiT~\cite{wu2024ptq4dit} & 8/8 & 0.1836 & 0.9991 & 54.56 & 53.33 & 0.440 \\
             % & SQ-Static~\cite{smooth_quant} & 8/8 & 0.1950 & 0.9980 & 55.2081 & 38.0752 & 0.216  \\
             % & ZeroQuant~\cite{zeroquant} & 8/8 & 0.1938 & 0.9988 & 58.3127 & 52.5630 & 0.032  \\   
             ViDiT-Q~\cite{zhao2024vidit} & 8/8 & 0.1950 & 0.9991 & 60.70 & 54.64 & 0.089  \\

             TR-DQ & 8/8 & 0.1861 &0.9990 & 62.43 & 57.07 & 0.295 \\
             TR-DQ+AS & 8/8 & 0.1830 & 0.9991 & 59.16 & 51.52 & 0.128 \\
            \cmidrule(lr){1-7}
              % & Naive PTQ & 6/6 & 0.1912 & 0.9964 & 14.6374 & 10.6919 & 32.636  \\
              % PTQ4DM & 6/6 & 0.1804 & 0.9977 & 20.9610 & 8.6566 & 0.2750  \\
              % Q-Diffusion & 6/6 & 0.1680 & 0.9963 & 15.0521 & 1.914 & 19.676 \\
              % PTQD & 6/6 & 0.1677 & 0.9962 & 7.998 & 1.4854 & 36.199 \\
              % PTQ4DM & 6/6 & - & - & - & - & -  \\
              % Q-Diffusion & 6/6 & - & - & - & - & - \\
              % PTQD & 6/6 & - & - & - & - & - \\
              Q-DiT~\cite{chen2024q} & 6/6 & 0.1710 & 0.9943 & 11.04 & 1.869 & 41.10 \\
              PTQ4DiT~\cite{wu2024ptq4dit} & 6/6 & 0.1799 & 0.9976 & 59.97 & 43.89 & 0.997 \\
              % & SQ-Static~\cite{smooth_quant} & 6/6 & 0.1792 & 0.9975 & 10.8512 & 9.4423 & 0.4160 \\
              % & ZeroQuant~\cite{zeroquant} & 6/6 & 0.1815 & 0.9962 & 8.218 & 6.3341 & 34.439 \\
          ViDiT-Q~\cite{zhao2024vidit} & 6/6 & 0.1791 & 0.9984 & 64.45 & 51.58 & 0.625 \\
             TR-DQ & 6/6 & 0.1795 & 0.9988 & 61.80 & 49.58 & 0.042 \\
             TR-DQ+AS & 6/6 & 0.1747 & 0.9987 & 59.74 & 44.94 & 0.068\\
             %  & ViDiT-Q-MP & 6/6 & 0.1794 & 0.9983 & 60.8124 & 53.4413 & 0.2516 \\
            \cmidrule(lr){1-7}
              % & Naive PTQ & 4/8 & 0.2010 & 0.9986 & 0.1765 & 0.0863 & 0.597  \\
              % PTQ4DM~\cite{ptq4dm} & 4/8 & 0.1727 & 0.9981 & 0.4781 & 0.2672 & 0.555  \\
              % & PTQ4DM + MP~\cite{ptq4dm} & 4/8 & 0.1808 & 0.9979 & 34.4329 & 24.8963 & 0.016  \\
              % & SQ-Static~\cite{smooth_quant} & 4/8 & 0.1950 & 0.9980 & 0.3671 & 1.4528 & 0.564  \\
              % PTQ4DM & 4/8 & - & - & - & - & -  \\
              % Q-Diffusion & 4/8 & - & - & - & - & - \\
              % PTQD & 4/8 & - & - & - & - & - \\
              Q-DiT~\cite{chen2024q} & 4/8 & 0.1687 & 0.9833 & 0.007 & 0.018 & 3.013 \\
              PTQ4DiT~\cite{wu2024ptq4dit} & 4/8 & 0.1735 & 0.9973 & 2.210 & 0.318 & 0.108 \\

              ViDiT-Q~\cite{zhao2024vidit} & 4/8 & 0.1809 & 0.9989 & 60.62 & 49.38 & 0.153  \\

              TR-DQ & 4/8 & 0.1815 & 0.9990 & 59.86 & 55.56 & 0.130 \\
              TR-DQ+AS & 4/8 & 0.1715 & 0.9993 & 56.87 & 48.09 & 0.306 \\
            \bottomrule[1pt]
            \vspace{-20pt}
            \end{tabular}}
\caption{\textbf{The corresponding effects of different quantization methods on prompt.} Video generated at different bit-widths with response to prompt. q-diffusion does not generate video properly at W6A6 and W4A8.}
    \label{fig:main_quant_comparison}
\end{table*}
\noindent\textbf{Baseline.}
Since many diffusion quantization methods are developed from LLMs quantization methods, our baseline includes diffusion quantization methods and LLMs quantization methods. Therefore, the quantization schemes for LLMs that we have chosen include SmoothQuant~\cite{xiao2023smoothquant}, DuQuant~\cite{lin2025duquant}, and Quarot~\cite{ashkboos2024quarot}, all of which can achieve weight-activation quantization. The selected diffusion quantization baselines include Q-Diffusion~\cite{li2023q}, Q-DiT~\cite{chen2024q}, PTQ4DiT~\cite{wu2025ptq4dit} and ViDiT-Q~\cite{zhao2024vidit}.

\noindent\textbf{Evaluation Metrics.}
Following ViDiTQ~\cite{zhao2024vidit}, for the image generation task, we chose FID~\cite{heusel2017gans} for fidelity evaluation, Clipscore for text-image alignment~\cite{hessel2021clipscore}, and ImageReward~\cite{xu2023imagereward} for human preference. For the video generation model, we choose Vbench benchmark~\cite{huang2024vbench} as the evaluation metric. For the model efficiency task, we utilize inference Peak Memory, and throughout of model. Also for the relationship between language and video, we chose follow EvalCrafter~\cite{liu2024evalcrafter}. we chose CLIPSIM and CLIP-Temp to measure text-video alignment and temporal semantic consistency, as well as DOVER's~\cite{wu2023exploring} Video Quality Assessment (VQA) metrics to assess the quality of the generation from both an aesthetic (high level) and technical (low level) perspective, Flow-score and Temporal Flickering for assessing temporal consistency.

As it shown in Tab.~\ref{tab:image} and Fig.~\ref{image}, the model quantinaized by TR-DQ leads the model quantinaized by the other quantitative methods in all metrics. At the same number of quantization bits, both TR-DQ and the method after weight sharing generate images with better quality than ViDiT-Q. This suggests that many of the previous methods ignored the time-steps change of MASSIVE activation and activation, while TR-DQ proposed a more effective solution. When the weights were quantinaized to 4bit, the effect of massive activation was ignored as ViDiT only smoothed the outliers of the weights by rotating the matrix. Therefore, in the case of diffusion quantization, the treatment of activations greatly affects the performance of the quantinaized  model. In contrast, our approach focuses on the activations, which are smoother and more favourable for quantization.

\begin{table*}[t]
\centering

\resizebox{0.85\linewidth}{!}{

\begin{tabular}{ccccccccccc}
\toprule
\multicolumn{5}{c}{\textbf{Methods}} & \textbf{Bit-width} & \multirow{2}{*}{\textbf{CLIPSIM}} & \multirow{2}{*}{\textbf{CLIP-Temp}} & VQA-      & VQA-      & $\Delta$ Flow \\ \cmidrule{1-6}
\textbf{Smooth}  & $\mathbf{R_{1} } $  & $\mathbf{P} $ & $\mathbf{R_{2} } $ & \textbf{T-R} & \textbf{(W/A)}     &                          &                            & Aesthetic & Technical & Score.        \\ \midrule
        -& - & - & - & - & 16/16 & 0.1797 & 0.9988 & 63.40 & 50.46 & - \\
        \ding{51} & - & - & - & - & 4/8 & 0.1739 & 0.9985 & 44.12 & 21.19 & 0.675 \\
        \ding{51} & \ding{51} & - & - & - & 4/8 & 0.1755 & 0.9941 & 50.42 & 42.12 & 0.421 \\
        \ding{51} & \ding{51} & - & \ding{51} & - & 4/8 & 0.1745 & 0.9972 & 52.38 & 45.67 & 0.342 \\
        \ding{51} & \ding{51} & \ding{51} & \ding{51} & - & 4/8 & 0.1741 & 0.9985 & 54.94 & 47.97 & 0.219 \\
        \ding{51} & \ding{51} & \ding{51} & \ding{51} & \ding{51} & 4/8 & 0.1815 & 0.9990 & 59.86 & 55.56 & 0.130 \\
  \bottomrule
\end{tabular}}
\vspace{-10pt}

\caption{\textbf{Ablation studies of TR-DQ.} We discuss the main influences on the model when quantifying W4A8.}
\label{ablation}
\end{table*}

\begin{table}[]
    \centering
    \resizebox{0.95\linewidth}{!}{
    \begin{tabular}{ccccc}
\toprule
\textbf{Bit-width} & \multicolumn{2}{c}{\textbf{A800}}  & \multicolumn{2}{c}{\textbf{A100}}  \\ 
\textbf{(W/A)}     & \textbf{Memory} & \textbf{Latency} & \textbf{Memory} & \textbf{Latency} \\ \midrule
16/16              & 1.00$\times$           & 1.00$\times$             &        1.00$\times$           & 1.00$\times$            \\ \midrule
8/8 (ViDiTQ)        & 1.98$\times$            & 1.70$\times$             &         2.00$\times$        &        1.74$\times$          \\
8/8 (TR-DQ)         & 1.97$\times$               &      1.69$\times$             &      1.97$\times$           &         1.69$\times$           \\ 
8/8 (TR-DQ+AS)         & 2.17$\times$               &    1.89$\times$               &       2.17$\times$          &         1.91$\times$         \\\midrule
4/8 (ViDiTQ)        & 2.41$\times$          & 1.36×            &    2.42$\times$             &       1.38$\times$           \\
4/8 (TR-DQ)         & 2.46$\times$              &       1.38$\times$           &    2.47$\times$             &         1.42$\times$         \\ 
4/8 (TR-DQ+AS)         & 2.58$\times$              &          1.41$\times$         &        2.59$\times$         &       1.44$\times$           \\\bottomrule
\end{tabular}}
    \caption{\textbf{Efficiency Comparison between original model and SOTA method.} The size and lantency of the compressed model of TR-DQ is almost the same as that of ViDiTQ.}
    \label{tab:efficient}
    \vspace{-1.5em}
\end{table}

\noindent \textbf{Video Generation Tasks.}
%\begin{figure}
%    \centering
%    \includegraphics[width=\linewidth]{ICCV2025-Author-Kit-Feb/sec/figure/long.pdf}
%    \caption{\textbf{Visualisation of long prompt.}}
%    \label{fig:enter-label}
%\end{figure}
The results of the experimental video generation evaluation may have some errors due to the poor robustness of the benchmark vbench. As it shown in Tab.~\ref{tab:main_vbench}, TR-DQ is better than the current SOTA ViDiTQ in most metrics. especially when the weights are quantized to 4bit. This difference is more significant compared to images, and we believe that one reason is that the video generation model has a larger number of parameters compared to the image generation model, so the effect of activation quantization is more significant. In addition, the video generation task has a stronger timing dependency than the image generation task, so other quantization methods may lack fine-grained timing divisions, leading to poorer quality of the generated video. As shown in Tab.~\ref{fig:main_quant_comparison}, for most metrics, TR-DQ is ahead of other methods. This suggests that the video-text consistency of the quantized TR-DQ model has an advantage over other methods. And this gap becomes more obvious when the weights are quantized to 4bit. This shows that we effectively shift the activation to the weights to reduce the error caused by the activation. Also, we propose a long prompt video generation sample to for visualisation. The video quality generated by the compressed model is slightly degraded compared to the original model, but still maintains better results. The comparison with the quantization approach to large language model can be found in the supplementary material.

\noindent \textbf{Classifier-Free Guidance Result.}
 As it shown in Tab.~\ref{tab:image}, with the addition of CFG weight sharing in TR-DQ, although there is a loss in the quality of the generated image, its effect still outperforms the current SOTA method. Therefore, it indicates that the difference between partial CFG and non-CFG attention distribution is not obvious, and there is a possibility of compression. As it shown in Tab.~\ref{tab:main_vbench} and Tab.~\ref{fig:main_quant_comparison},  Attention Sharing leads to a decrease in the quality of video generation, but the generative power of the model still differs little from the current SOTA. Therefore, our approach reduces redundant ATTENTION computations while maintaining model generation capabilities.

\begin{figure}[t]
    \centering
    \begin{subfigure}{0.48\linewidth}
        \centering
        \includegraphics[width=\linewidth]{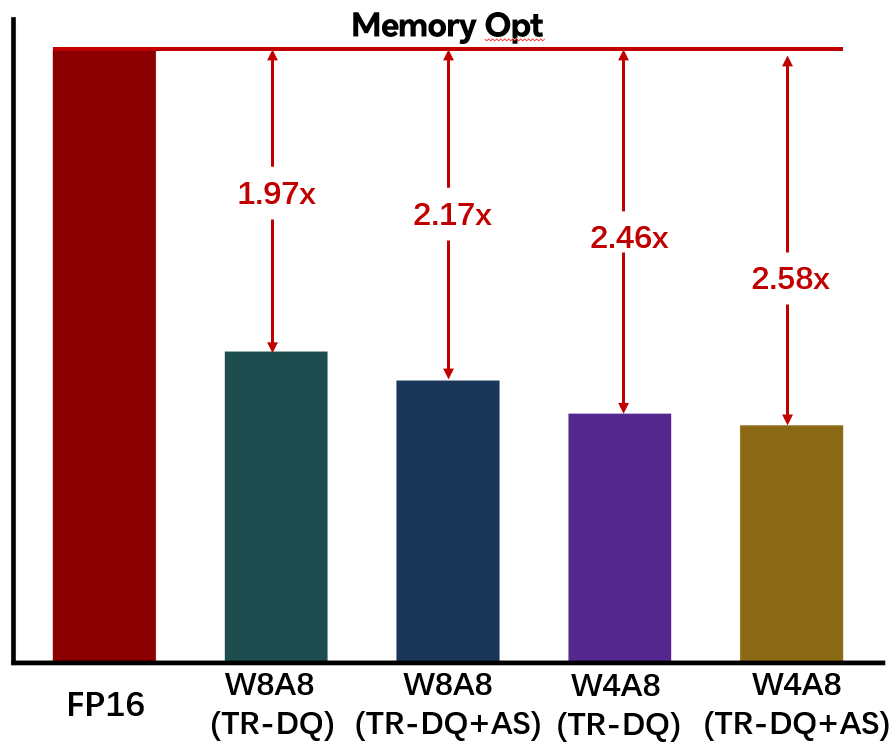}
    \end{subfigure}
    \hfill
    \begin{subfigure}{0.48\linewidth}
        \centering
        \includegraphics[width=\linewidth]{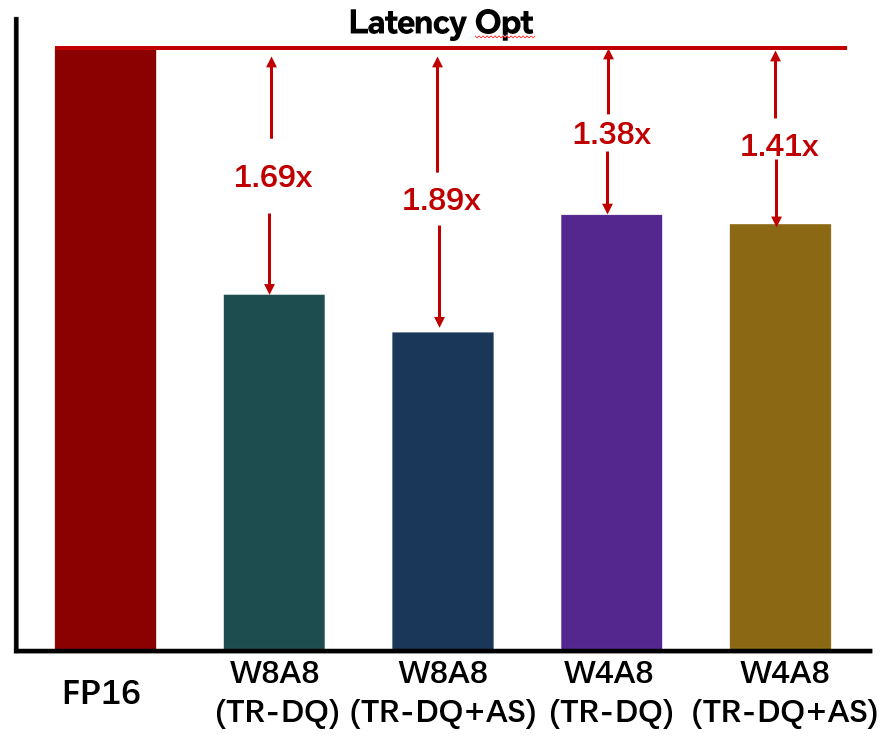}
    \end{subfigure}
    \caption{\textbf{Further Efficiency Visualization of TR-DQ.} We enhance the clarity of our approach's high efficiency through further visualization.}
    \label{efficient}
\end{figure}

\noindent \textbf{Efficiency Comparison.}
 As it shown in Tab.~\ref{tab:efficient} and Fig.~\ref{efficient}, TR-DQ significantly reduces the memory overhead and lantency compared to the original model, and is more conducive to hardware inference computation because the TR-DQ activation and weight distributions are smoother than those of ViDiTQ. Further, we reduce the attention computation by weight sharing so that the model can skip the layers with high CFG, no-CFG similarity in inference. This operation reduces the overall inference overhead of the model and also reduces the computational latency of the model.

\subsection{Ablation Study.}

In this section, we discuss the main influences of our methodology. We will discuss our main contribution from the rotation matrix approach and time modeling.

As it shown in Tab.~\ref{ablation}, $\mathbf{R_{1} } $ is the rotation matrix for handling activation outliers, $\mathbf{R_{2} } $ is the rotation matrix for handling weight outliers, and $\mathbf{P }  $ denotes the weight permutation process. We found that the biggest impact on the model was the dynamic transformation of the rotation matrix based on time-steps. The overall quality of the generated video is significantly improved by adding time information. The fact that $\mathbf{R_{1} } $ has a greater impact on the overall effect of the video than any other factor suggests that diffusion is different from the large language model in that it directly affects the quality of the generation of the generative model. Permutation and $\mathbf{R_{2} } $, although both affect the video generation results, are not major factors. In contrast $\mathbf{R_{2} } $ has a greater impact than permutation.

\section{Conclusion}
In this article, we explore two current issues in diffusion model quantization: massive activation and time-steps sampling. To address these problems, we design a rotation matrix quantization method based on time-steps activation distribution,~\textbf{Time-Rotation Diffusion Quantization (TR-DQ)}. TR-DQ shifts hard to quant activations to weights via a matrix and adaptively adjusts the parameters of the rotation matrix for each time-step activation change. Meanwhile, we found that some layers have higher weight similarity in the case of CFG and non-CFG, so we chose to merge these weights for processing to reduce the memory overhead of CFG. Our method has better image generation and video generation and generates results compared to current quantization methods. Compared to original model, our approach achieve 1.38-1.89$\times$ speedup and 1.97-2.58$\times$ memory reduction.

{
    \small
    \bibliographystyle{ieeenat_fullname}
    \bibliography{main}
}
\clearpage
\setcounter{page}{1}
\maketitlesupplementary

\section{More Result}
\label{Result}
As it shown in Tab.~\ref{tab:main_vbench1} and Tab.~\ref{tab:main_quant_comparison}, comparing to the llm quantization method, our method has a significant improvement. Meanwhile, DuQuant generates videos that are basically similar to ViDiTQ, thus indicating that the model activation quantization method has a large impact on the final results.

\begin{table*}[htb]

\centering
    \begin{tabular}{ccccccccccc}
\toprule
\multirow{2}{*}{\textbf{Method}} & \textbf{Bit-width} & Imaging & Aesthetic & Motion & Dynamic & BG. & Subject & Scene & Overall \\
& (W/A) & Quality & Quality & Smooth. & Degree &  Consist. & Consist. & Consist. & Consist. \\ 
\midrule 
 - & 16/16 & 63.68 & 57.12 & 97.01 & 56.94 & 96.13 & 92.28 & 40.51 & 26.21 \\
\midrule 
 % Naive PTQ & 8/8 & 56.52 & 53.64 & 95.85 & 69.44 & 93.54 & 85.30 & 29.28 & 25.42 &   \\

 % SQ-Static & 8/8 & 61.74 & 55.93 & 95.79 & 66.66 & 95.18 & 88.70     & 35.73 & 26.31 &   \\
 SmoothQuant & 8/8 & 62.22 & 55.90 & 95.96 & 68.05 & 94.17 & 87.71 & 36.66 & 25.66 &   \\
 Quarot & 8/8 & 60.14 & 53.21 & 94.98 & 66.21 & 95.03 & 85.35 & 35.65 & 25.43 &   \\

 DuQuant & 8/8 & 60.44 & 56.97 & 95.76 & 50.12 & 96.67 & 91.45 & 34.67 & 25.95 &   \\
 TR-DQ & 8/8 & 61.82 & 57.44 & 96.63 & 55.38 & 96.11 & 91.14 & 39.78 & 26.18    \\
 TR-DQ+AS & 8/8 & 60.38 & 57.10 & 96.26 & 50.27 & 95.71 & 91.58 & 38.50 & 25.99    \\
\midrule 
  % & ViDiT-Q-MP & 6/6 & 62.07 & 57.03 & 95.86 & 62.50 & 95.86 & 89.34 & 39.46 & 26.41   \\

 SmoothQuant & 4/8 & 46.98 & 44.38 & 94.59 & 21.67 & 94.36 & 82.79 & 26.41 & 18.25 &   \\
 Quarot & 4/8 & 44.25 & 43.78 & 92.57 & 66.21 & 94.25 & 84.55 & 28.43 & 18.43 &  \\
 % ViDiT-Q & 4/8 & 49.25 & 45.48 & 98.20 & 27.77 & 96.91 & 91.46 & 25.41 & 20.42 \\

 DuQuant & 4/8 & 58.11 & 53.29 & 95.32 & 47.56 & 96.61 & 92.19 & 31.37 & 24.29 &   \\
TR-DQ & 4/8 & 59.88 & 56.20 & 96.57 & 51.83 & 96.65 & 90.74 & 32.46 & 26.17   \\
TR-DQ+AS & 4/8 & 57.69 & 55.02 & 96.74 & 47.78 & 96.58 & 91.03 & 32.25 & 25.34    \\
\bottomrule
\vspace{-20pt}
\end{tabular}
\caption{\textbf{Performance of TR-DQ on video generation on VBench.} TR-DQ outperforms the current LLM quantization metrics, suggesting that it is more capable of generating models after quantization.}
\label{tab:main_vbench1}
\end{table*}

\begin{table*}[htb]
   \centering
        \resizebox{0.75\linewidth}{!}{
            \begin{tabular}{ccccccc}
            \toprule
            % \multirow{2}{*}{\textbf{Model}} & 
            \multirow{2}{*}{\textbf{Method}} & \textbf{Bit-width} & \multirow{2}{*}{\textbf{CLIPSIM}} &  \multirow{2}{*}{\textbf{CLIP-Temp}} & VQA- & VQA- & $\Delta$ Flow  \\
             & (W/A) & & & \small{Aesthetic} & \small{Technical} & Score. ($\downarrow$) \\
            \midrule
            
            % \multirow{24}{*}{\textbf{\large STDiT}} & 
            - & 16/16 &  0.1818 & 0.9988 & 63.40 & 50.46 & - \\
             \cmidrule(lr){1-7}
             % & Naive PTQ & 8/8 & 0.1956 & 0.9988 & 48.2358 & 25.5961 & 0.367  \\
             % & SQ-Static~\cite{smooth_quant} & 8/8 & 0.1950 & 0.9980 & 55.2081 & 38.0752 & 0.216  \\
             % & ZeroQuant~\cite{zeroquant} & 8/8 & 0.1938 & 0.9988 & 58.3127 & 52.5630 & 0.032  \\
             SmoothQuant & 8/8 & 0.1951 & 0.9986 & 59.78 & 51.53 & 0.331  \\
             Quarot & 8/8 & 0.1949 & 0.9976 & 58.73 & 52.28 & 0.215  \\
             
             Duquant & 8/8 & 0.1745 & 0.9988 & 56.35 & 55.66 & 0.161 \\
             TR-DQ & 8/8 & 0.1861 &0.9990 & 62.43 & 57.07 & 0.295 \\
             TR-DQ+AS & 8/8 & 0.1830 & 0.9991 & 59.16 & 51.52 & 0.128 \\
            \cmidrule(lr){1-7}
              % & Naive PTQ & 6/6 & 0.1912 & 0.9964 & 14.6374 & 10.6919 & 32.636  \\
              % PTQ4DM & 6/6 & 0.1804 & 0.9977 & 20.9610 & 8.6566 & 0.2750  \\
              % Q-Diffusion & 6/6 & 0.1680 & 0.9963 & 15.0521 & 1.914 & 19.676 \\
              % PTQD & 6/6 & 0.1677 & 0.9962 & 7.998 & 1.4854 & 36.199 \\
              % PTQ4DM & 6/6 & - & - & - & - & -  \\
              % Q-Diffusion & 6/6 & - & - & - & - & - \\
 
              SmoothQuant & 6/6 & 0.1807 & 0.9985 & 56.45 & 48.21 & 29.26 \\
              Quarot & 6/6 & 0.1820 & 0.9975 & 61.47 & 53.06 & 0.146  \\
            
              Duquant & 6/6 & 0.1761 & 0.9990 & 54.20 & 46.25 & 0.103 \\
             TR-DQ & 6/6 & 0.1795 & 0.9988 & 61.80 & 49.58 & 0.042 \\
             TR-DQ+AS & 6/6 & 0.1747 & 0.9987 & 59.74 & 44.94 & 0.068\\
             %  & ViDiT-Q-MP & 6/6 & 0.1794 & 0.9983 & 60.8124 & 53.4413 & 0.2516 \\
            \cmidrule(lr){1-7}
              % & Naive PTQ & 4/8 & 0.2010 & 0.9986 & 0.1765 & 0.0863 & 0.597  \\
              % PTQ4DM~\cite{ptq4dm} & 4/8 & 0.1727 & 0.9981 & 0.4781 & 0.2672 & 0.555  \\
              % & PTQ4DM + MP~\cite{ptq4dm} & 4/8 & 0.1808 & 0.9979 & 34.4329 & 24.8963 & 0.016  \\
              % & SQ-Static~\cite{smooth_quant} & 4/8 & 0.1950 & 0.9980 & 0.3671 & 1.4528 & 0.564  \\
              % PTQ4DM & 4/8 & - & - & - & - & -  \\
              % Q-Diffusion & 4/8 & - & - & - & - & - \\
              % PTQD & 4/8 & - & - & - & - & - \\

              SmoothQuant & 4/8 & 0.1832 & 0.9983 & 31.96 & 22.85 & 0.415\\
              Quarot & 4/8 & 0.1817 & 0.9965 & 47.36 & 33.13 & 0.326  \\
              % ViDiT-Q & 4/8 & 0.1812 & 0.9989 & 60.2159 & 42.2571 & 0.151\\
              Duquant & 4/8 & 0.1741 & 0.9985 & 54.94 & 47.97 & 0.219 \\
              TR-DQ & 4/8 & 0.1815 & 0.9990 & 59.86 & 55.56 & 0.130 \\
              TR-DQ+AS & 4/8 & 0.1715 & 0.9993 & 56.87 & 48.09 & 0.306 \\
            \bottomrule[1pt]
            \vspace{-20pt}
            \end{tabular}}
\caption{\textbf{The corresponding effects of different quantization methods on prompt.} The comparison of generation quality for different quantization methods under different bit-widths.}
    \label{tab:main_quant_comparison}
\end{table*}

%\clearpage

\end{document}